\documentclass[a4, 10 pt, conference]{ieeeconf}  

\IEEEoverridecommandlockouts                              

\newcommand{\customfootnote}[1]{
    \begingroup
    \renewcommand{\thefootnote}{}
    \footnote{#1}
    \addtocounter{footnote}{-1}
    \endgroup
}

\usepackage{url}
\title{\LARGE \bf
A Summarized History-based Dialogue System \\ for Amnesia-Free Prompt Updates
}

\author{Hyejin Hong$^{1}$, Hibiki Kawano$^{1}$, Takuto Maekawa$^{1}$, Naoki Yoshimaru$^{2}$, Takamasa Iio$^{1}$ and Kenji Hatano$^{1}$
\thanks{*This work was partially supported by JSPS KAKENHI Grant Number 19H05691, 22H03895, 23H03694 and the Grants-in-Aid for Academic Promotion, Graduate School of Culture and Information Science, Doshisha University.}
\thanks{$^{1}$ Faculty of Culture and Information Science, Doshisha University, 1-3 Tatara-Miyakodani
Kyotanabe, Kyoto 610-0394, Japan}%
\thanks{$^{2}$ Graduate School of Culture and Information Science, Doshisha University, 1-3 Tatara-Miyakodani
Kyotanabe, Kyoto 610-0394, Japan}%
}

\usepackage{graphicx}
\graphicspath{ {./images/} }
\begin{document}

\maketitle
\thispagestyle{empty}
\pagestyle{empty}

\begin{abstract}
In today's society, information overload presents challenges in providing optimal recommendations. 
Consequently, the importance of dialogue systems that can discern and provide the necessary information through dialogue is increasingly recognized. 
However, some concerns existing dialogue systems rely on pre-trained models and need help to cope with real-time or insufficient information. 
To address these concerns, models that allow the addition of missing information to dialogue robots are being proposed. 
Yet, maintaining the integrity of previous conversation history while integrating new data remains a formidable challenge. 
This paper presents a novel system for dialogue robots designed to remember user-specific characteristics by retaining past conversation history even as new information is added.
\end{abstract}

\section{Introduction}
\vspace{-0.1cm}
Today’s society faces the problem of information overload. 
For example, travel agencies face the challenge of recommending travel destinations based on customers’ experiences, but with so much information worldwide, it is difficult to remember it all\cite{DRC2023}. 
To solve this problem, information recommendation using dialogue systems is becoming increasingly important\cite{DialogueSystems}. 
This is because dialogue systems enable targeted information gathering through contextual questions and information recommendations tailored to each customer.

Moreover, recent advancements in dialogue technologies with the Large Language Models (LLMs) have significantly enhanced dialogue performance. 
Due to their reliance on pre-trained models, these models exhibit a significant challenge in their ability to integrate real-time, up-to-date, and comprehensive information\cite{Yong2023PromptingML}. 
Conventional methodologies, therefore, have been trying to update the robot’s profile from information obtained from the user during the conversation. 
However, existing dialogue systems have identified a significant limitation: the inability to reference previous conversation data when profiles are modified during an ongoing dialogue. 
Therefore, our system incorporates a dual-storage approach to maintain uninterrupted conversations, even with updates to the robot’s role. 
Our system features storage for current dialogues and another for summaries, with the latter used for updating the robot’s role.

\begin{figure}[t]
\centering
    \includegraphics[width=8.5cm]{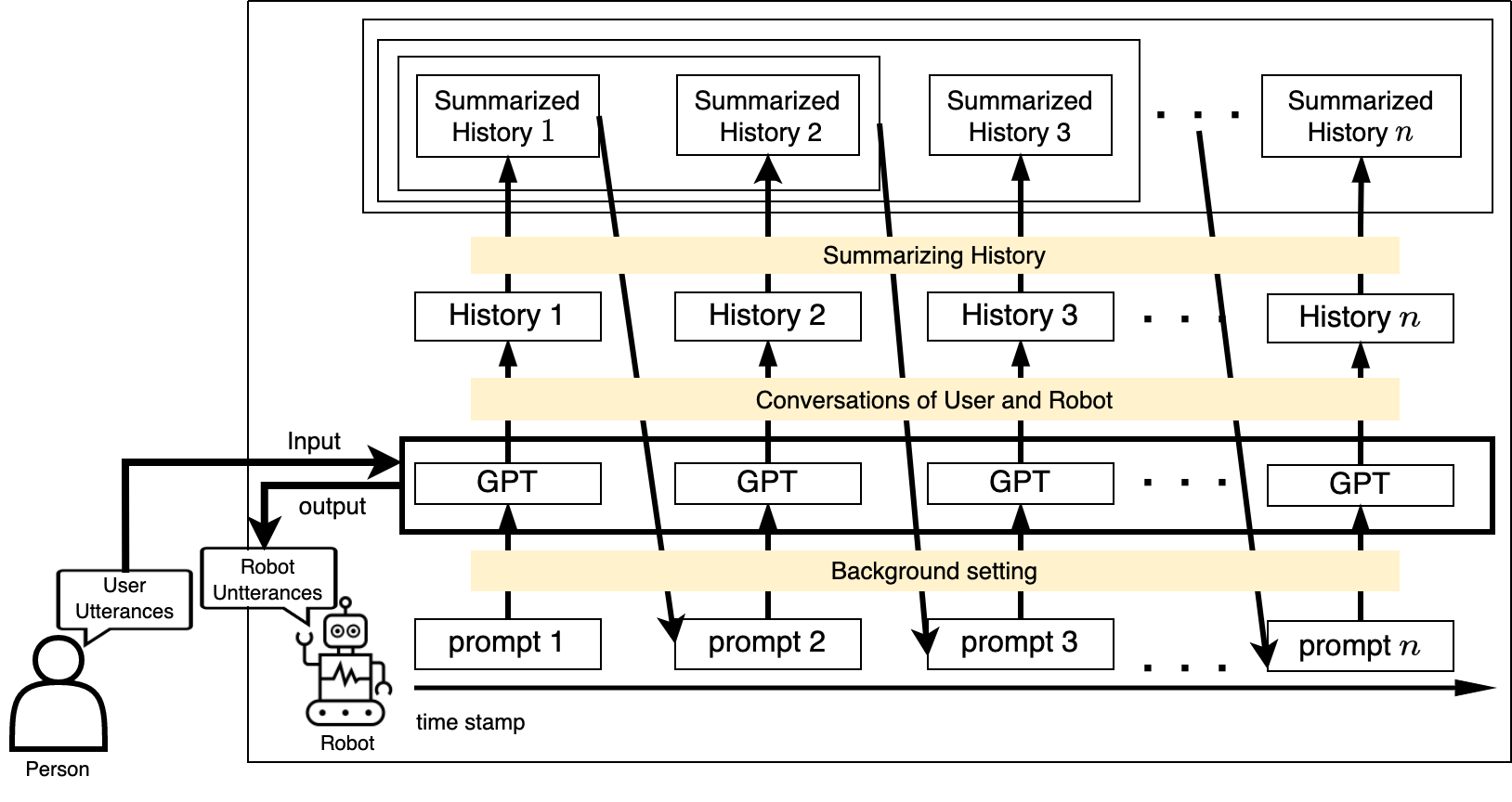}
    \vspace{-0.2cm}
    \caption{\textbf{A Summarized History-based Dialogue System}: This figure shows how the dialogue system maintains conversation history and updates prompts. 
    The system consistently summarizes and integrates past dialogues to form new prompts, retaining all details from previous conversations. 
    The variable $n$ indicates how many times the prompt has been updated.
    }
    \
    \label{fig:urelabel}
    \vspace{-0.6cm}
\end{figure}

\section{Proposed Method}
\vspace{-0.1cm}
We propose a system capable of adding new data and information without losing memory, even when the background or prompt is updated during the conversation. 
This system aims to enhance the integration of new information and the retention of dialogue content.

However, continuously storing history can cause memory overflow, so we propose a system that stores a summary of the history\cite{Wang2023RecursivelySE}.
Essentially, there are two history storage components: one for storing the history of conversations with the current prompt and another for storing summarized histories of conversations with previous prompts. 
This dual-storage approach prevents the internal memory from becoming overly large.
In Fig. \ref{fig:urelabel}, within the Integrated of History, there is a possibility of memory overflow as the number of conversation turns increases. 
In this study, we propose a method where, when the prompt is updated, the current conversation content is combined with previously summarized material to create a newly summarized history. 
This approach ensures that both recent interactions and past summaries are seamlessly integrated into a cohesive and updated historical record.
In essence,  prevent memory overflow, we uti- lize LLMs to iteratively generate summaries.

Our proposed system is illustrated in Fig. \ref{fig:urelabel}.
The overall flow of the dialogue system is as follows:

\begin{LaTeXenumerate}
\item Enter the prompts that establish the robot's background into the system.
\item Activate the dialogue system when the user initiates communication with the robot.
\item Input the user's utterances into the GPT4\cite{openai2023gpt4} model, which then produces the robot's responses. 
\item After a set amount of time has elapsed, summarize the history $n$ and integrate it with the previous summarized history.
\item Use this combined history to update the background prompt.
\item Repeat steps 2 to 5 in a continuous loop.
\end{LaTeXenumerate}

\vspace{-0.1cm}
\section{Experiment}
\vspace{-0.1cm}
The evaluation of the proposed system was conducted within a travel agency context, where the robot assumed the role of a consultant, assisting in developing travel plans instead of human staff. 

The assessment comprised two primary components: a specific evaluation of the travel plans formulated and a nine-item impression evaluation.

First, regarding the travel plans, an evaluation was conducted on whether a feasible plan had been created for visiting two tourist spots.
The nine-item impression evaluation was conducted on the following twelve questions:

\begin{LaTeXenumerate}
\renewcommand{\labelenumi}{Q\arabic{enumi}:}
\setlength{\leftskip}{1em}
\item Could you sufficiently obtain information about the tourist spots?
\item Could you have a natural dialogue with the robot?
\item Was the robot’s response appropriate?
\item Was the robot’s response favorable?
\item Were you satisfied with your dialogue with the robot?
\item Did you find the robot to be trustworthy?
\item Did you refer to the information provided by the robot when choosing tourist spots?
\item Did you think the information provided by the robot was accurate?
\item Would you like to use this robot’s service again?
\item Could you make a plan for visiting two tourist sites?
\item Did you think the plan you have prepared (based on your own common sense) is feasible?
\item Did you have a plan to visit the two destinations and did it seem feasible?
\end{LaTeXenumerate}

The results indicated that the system successfully achieved high satisfaction levels across both the baseline impression and plan assessments.
However, among the nine-item impression evaluation, the scores for Q2 and Q8 were lower than the baseline.

\def\tablename{Table}
\begin{table}[t]
\vspace{-0.2cm}
\centering
\caption{Average of evaluation results}
\label{table_example}
\vspace{-0.3cm}
\begin{tabular}{c|r|r}
\hline
Team & Impression Assessment & Plan \\\hline
\textbf{Jerry-cis} & \textbf{4.476} & \textbf{0.857} \\
Baseline & 4.298 & 0.77 \\
\hline
\end{tabular}
\vspace{-0.5cm}
\end{table}

\vspace{-0.1cm}
\section{Discussion}
\vspace{-0.1cm}
In the conducted evaluation, it was observed that the overall scores surpassed those of the baseline. 
This observation can be attributed to the continuous updating of prompts, enabling the reflection of user characteristics during the dialogue. 
Moreover, the utilization of dual-strage to store summaries of past interactions contributed to preserving user traits, enhancing personalized dialogues as the number of prompt updates grew. 
Nonetheless, in the assessment of the nine-item impression evaluation, Q2, Q8 scored lower than the baseline. 
This underperformance is conjectured to stem from the need for more incorporation of robot interaction elements in the study design.

The proposed system predominantly emphasized the internal dynamics of conversation flow. 
However, user satisfaction with real-world dialogue robots is not solely dependent on the content or progression of dialogue. Still, it is also significantly influenced by external factors such as the robot's speech modulation, movements, and gaze direction \cite{de2017influence}. 
The credibility of the dialogue content varies with the robot's manner of speech, motion, and gaze direction. 
However, during the development phase, extensive robot behaviors were deliberately not implemented to avoid diminishing user satisfaction due to potentially unnatural robot movements and gaze \cite{uncanny}. 
This restraint likely contributed to the need to establish trust towards the robot.
Additionally, we think flaws in trust formation with the dialogue robot have diminished trust in its information.

\section{Conclusion}
\vspace{-0.1cm}
In this study, we proposed a system that can update the background, namely the prompt, during a conversation without losing memory and can add new data and information. 
In an evaluation against the baseline system, the proposed system showed higher user satisfaction with the dialogue system used by the robot in most of the items.

Future tasks include refining robot interactions to bolster user trust in dialogue robots. 
At this stage, it's important to take into account the Uncanny Valley effect. 
In this iteration, we also implemented a feature that pauses the robot's speech and listens to the user if they interrupt. 
However, this feature became impractical due to the robot mistaking ambient noise for user speech. 
Therefore, it is necessary to consider methods to distinguish between background noise and user speech, enabling the development of a system that can more effectively utilize user utterances to gather information.

\addtolength{\textheight}{-12cm}  


\customfootnote{
AUTHOR CONTRIBUTION:
Conceptualization, H.H; 
Software, H.H, H.K, T.M and N.Y; 
Investigation, H.H, H.K and T.M; 
Writing – Original, H.H;
Writing – Editing, N.Y, T.I and K.H;
Supervision T.I and K.H. 
}


\begin{thebibliography}{99}

\bibitem{DRC2023}
T. Minato, R. Higashinaka, K. Sakai, T. Funayama, H. Nishizaki, and T. Nagai, "Overview of Dialogue Robot Competition 2023," \emph{Proceedings of the Dialogue Robot Competition 2023}, 2023.

\bibitem{DialogueSystems}
H. Chen, X. Liu, D. Yin, and J. Tang, "A Survey on Dialogue Systems: Recent Advances and New Frontiers," \emph{ACM SIGKDD Explorations Newsletter}, vol. 19, no. 2, pp. 25-35, Nov. 2017.

\bibitem{Yong2023PromptingML}
Z.-X. Yong, R. Zhang, J. Z. Forde, S. Wang, A. Subramonian, S. Cahyawijaya, H. Lovenia, G. I. Winata, L. Sutawika, J. C. B. Cruz, L. Phan, Y. Tan, A. F. Aji, "Prompting Multilingual Large Language Models to Generate Code-Mixed Texts: The Case of South East Asian Languages," \emph{ArXiv}, 2023, [Online]. Available: \url{https://api.semanticscholar.org/CorpusID:257757220}.

\bibitem{Wang2023RecursivelySE}
Q. Wang, L. Ding, Y. Cao, Z. Tian, S. Wang, D. Tao, and L. Guo, "Recursively Summarizing Enables Long-Term Dialogue Memory in Large Language Models," \emph{ArXiv}, 2023, [Online]. Available: \url{https://api.semanticscholar.org/CorpusID:261276822}.

\bibitem{openai2023gpt4}
OpenAI, "GPT-4 Technical Report,"\emph{ArXiv}, 2023, [Online]. Available: \url{https://arxiv.org/abs/2303.08774}.

\bibitem{de2017influence}
M. M. A. de Graaf and S. B. Allouch, "The influence of prior expectations of a robot’s lifelikeness on users’ intentions to treat a zoomorphic robot as a companion," \emph{International Journal of Social Robotics}, vol. 9, pp. 17-32, 2017.

\bibitem{uncanny}
M. Mori, K. F. MacDorman, and N. Kageki, "The Uncanny Valley [From the Field]," \emph{IEEE Robotics \& Automation Magazine}, vol. 19, no. 2, pp. 98-100, 2012.

\end{thebibliography}
\end{document}